\documentclass{article} 
\usepackage[nonatbib,final]{neurips_2024}
\usepackage[numbers]{natbib} 
\usepackage{color-edits}
\usepackage{float}
\usepackage{upgreek}
\usepackage{amsbsy}
\usepackage{amssymb}
\usepackage{amsmath}
\usepackage{amsthm}
\usepackage{graphicx}
\usepackage{setspace}
\usepackage{mathtools}
\usepackage{xcolor}
\usepackage{nicefrac}
\usepackage{titlesec}
\usepackage{enumitem}
\usepackage{url}
\usepackage{caption}
\usepackage{subcaption}
\usepackage{adjustbox,booktabs,multirow}
\usepackage[breaklinks]{hyperref}
\usepackage[noabbrev]{cleveref}
\usepackage{autonum}
\usepackage{algorithm,algorithmic}
\usepackage{color-edits}
\hypersetup{
    colorlinks,
    linkcolor={red!100!black},
    citecolor={blue!100!black},
    urlcolor={blue!80!black}
}

\addauthor{vahid}{red}
\addauthor{keertana}{blue}
\addauthor{ethan}{green}
\addauthor{allison}{orange}

\makeatletter
\newtheorem*{rep@theorem}{\rep@title}
\newcommand{\newreptheorem}[2]{%
\newenvironment{rep#1}[1]{%
 \def\rep@title{#2 \ref{##1}}%
 \begin{rep@theorem}}%
 {\end{rep@theorem}}}
\makeatother

\newreptheorem{theorem}{Theorem}

\newreptheorem{corollary}{Corollary}

\newreptheorem{proposition}{Proposition}

\theoremstyle{definition}

\crefname{equation}{}{}
\crefname{proposition}{Proposition}{Propositions}
\crefname{appendix}{Appendix}{Appendices}
\crefname{lemma}{Lemma}{Lemmas}
\crefname{assumption}{Assumption}{Assumptions}
\crefname{algorithm}{Algorithm}{Algorithms}
\crefname{theorem}{Theorem}{Theorems}
\crefname{corollary}{Corollary}{Corollaries}
\crefname{definition}{Definition}{Definitions}
\crefname{figure}{Figure}{Figures}
\crefname{table}{Table}{Tables}

\DeclareMathOperator{\E}{\mathbb{E}}

\DeclareMathOperator{\R}{\mathbb{R}}  

\let\P\undefined

\DeclareMathOperator{\P}{\bbP}

\def\ddefloop#1{\ifx\ddefloop#1\else\ddef{#1}\expandafter\ddefloop\fi}
\def\ddef#1{\expandafter\def\csname bb#1\endcsname{\ensuremath{\mathbb{#1}}}}
\ddefloop ABCDEFGHIJKLMNOPQRSTUVWXYZ\ddefloop
\def\ddefloop#1{\ifx\ddefloop#1\else\ddef{#1}\expandafter\ddefloop\fi}
\def\ddef#1{\expandafter\def\csname b#1\endcsname{\ensuremath{\mathbf{#1}}}}
\ddefloop ABCDEFGHIJKLMNOPQRSTUVWXYZ\ddefloop
\def\ddef#1{\expandafter\def\csname c#1\endcsname{\ensuremath{\mathcal{#1}}}}
\ddefloop ABCDEFGHIJKLMNOPQRSTUVWXYZ\ddefloop
\def\ddef#1{\expandafter\def\csname h#1\endcsname{\ensuremath{\widehat{#1}}}}
\ddefloop ABCDEFGHIJKLMNOPQRSTUVWXYZ\ddefloop
\def\ddef#1{\expandafter\def\csname hat#1\endcsname{\ensuremath{\widehat{#1}}}}
\ddefloop abcdefghijklmnopqrstuvwxyz\ddefloop
\def\ddef#1{\expandafter\def\csname hc#1\endcsname{\ensuremath{\widehat{\mathcal{#1}}}}}
\ddefloop ABCDEFGHIJKLMNOPQRSTUVWXYZ\ddefloop
\def\ddef#1{\expandafter\def\csname t#1\endcsname{\ensuremath{\widetilde{#1}}}}
\ddefloop ABCDEFGHIJKLMNOPQRSTUVWXYZ\ddefloop
\def\ddef#1{\expandafter\def\csname tc#1\endcsname{\ensuremath{\widetilde{\mathcal{#1}}}}}
\ddefloop ABCDEFGHIJKLMNOPQRSTUVWXYZ\ddefloop

\usepackage{prettyref}
\usepackage{xcolor}

\newcommand{\midsem}{\,;\,}

\newcommand{\kl}[2]{\mathrm{KL}\left(#1 \ \middle\| \ #2\right)}

\newcommand{\data}{{D}}

\newcommand{\given}{\ \middle|\ }

\newcommand{\lpar}{\!\left(}
\newcommand{\rpar}{\right)}
\newcommand{\lbar}{\!\left[}
\newcommand{\rbar}{\right]}
\newcommand{\lcbar}{\!\left\{}
\newcommand{\rcbar}{\right\}}

\newcommand{\xhdr}[1]{\textbf{#1.}}

\newcommand{\pre}{\text{pre}}
\newcommand{\history}{H}
\newcommand{\dist}{\cD}
\newcommand{\loss}{\cL}

\title{Personalized Adaptation via In-Context Preference Learning }
\author{
  {Allison Lau}\\
  University of Toronto \\
  \texttt{allison.lau@mail.utoronto.ca}
  \And 
Younwoo (Ethan) Choi$^\star$ \\
University of Toronto \\
\texttt{ywchoi@cs.toronto.edu} 
\And 
Vahid Balazadeh$^\star$ \\
University of Toronto \\
\texttt{vahid@cs.toronto.edu} 
\And
Keertana Chidambaram\thanks{Equal contribution with random ordering.}\\
Stanford University \\
\texttt{vck@stanford.edu} 
\And
Vasilis Syrgkanis \\
Stanford University \\
\texttt{vsyrgk@stanford.edu}
\And
Rahul G. Krishnan\\
University of Toronto \\
\texttt{rahulgk@cs.toronto.edu} 
}

\newcommand{\method}{Preference Pretrained Transformer}
\newcommand{\methodabbrv}{PPT}

\begin{document}
\maketitle

\begin{abstract}
Reinforcement Learning from Human Feedback (RLHF) is widely used to align Language Models (LMs) with human preferences. However, existing approaches often neglect individual user preferences, leading to suboptimal personalization. We present the \method\ (\methodabbrv), a novel approach for adaptive personalization using online user feedback. \methodabbrv\ leverages the in-context learning capabilities of transformers to dynamically adapt to individual preferences. Our approach consists of two phases: (1) an offline phase where we train a single policy model using a history-dependent loss function, and (2) an online phase where the model adapts to user preferences through in-context learning. We demonstrate \methodabbrv's effectiveness in a contextual bandit setting, showing that it achieves personalized adaptation superior to existing methods while significantly reducing the computational costs. Our results suggest the potential of in-context learning for scalable and efficient personalization in large language models.
\end{abstract}

\section{Introduction}

Reinforcement Learning from Human Feedback (RLHF) has emerged as a powerful technique for aligning large language models (LLMs) with human preferences, enabling them to generate higher-quality and more desirable outputs~\citep{stiennon2020learning,nakano2021webgpt,bai2022training,ouyang2022training}. However, standard RLHF aims to learn a general policy optimized for the entire population, often neglecting the diversity of individual preferences and potentially marginalizing specific groups~\citep{vamplew2018human,rame2023rewardedsoupsparetooptimalalignment,jang2023personalizedsoupspersonalizedlarge,casper2023open,chakraborty2024maxminrlhfequitablealignmentlarge}. This limitation can lead to suboptimal experiences for users whose preferences deviate from the majority.
 
 Several approaches have been proposed to account for the diverse preferences of different populations. One strategy is to use multi-objective reinforcement learning (MORL)~\citep{barrett2008learning,li2020deep} techniques to train multiple (potentially interpolated) proxy rewards and their corresponding optimal policies during the training phase. Then, they choose the policy that maximizes each new user's reward as it becomes known during the selection phase~\citep{marta2023aligning,wu2024fine}. However, as mentioned in \citet{rame2023rewardedsoupsparetooptimalalignment}, those approaches require maintaining a large set of networks, potentially one for each possible preference profile. To address this issue, ~\citet{rame2023rewardedsoupsparetooptimalalignment} and \citet{jang2023personalizedsoupspersonalizedlarge} propose "rewarded soups" and "personalized soups," respectively, which learn separate reward and policy models for each preference criterion and aggregate the learned rewards or policies for new users. While effective, these approaches can be computationally expensive, particularly when the number of criteria is large, and may require re-training an entire model for every new criterion. Moreover, these methods often rely on a separate selection phase to identify each user's relevant reward/policy model (e.g., by optimizing the interpolating coefficients of the trained models), which translates to an extra computation step for every new user. 

\begin{figure}[t]
    \centering
    \includegraphics[width=\linewidth]{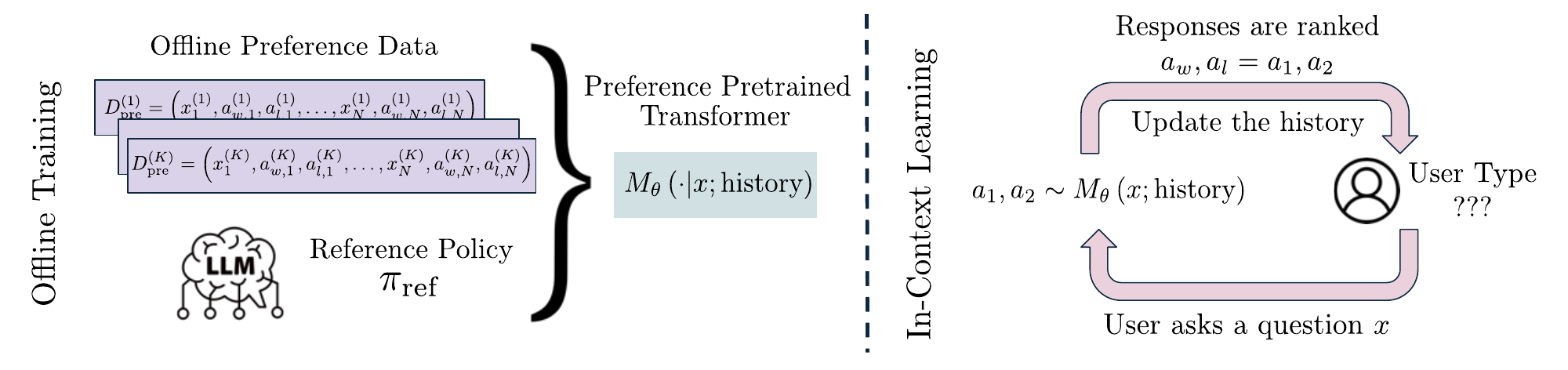}
    \caption{\method: (i) In the offline phase, we train a single policy to predict the preferred answers given the history of previous responses. (ii) In the online phase, the pretrained model interacts with the user, appends the interaction history to its context and generates more personalized responses.}
    \label{fig:main-figure}
\end{figure}

To address these challenges, we introduce a novel framework for online personalized adaptation without the need for training separate models of each preference criterion. Our intuition is that a \emph{history-dependent} policy should be able to identify the preference profile of new users after a few interactions with them. To learn such history-dependent policies, we leverage the in-context learning capabilities of transformer architectures~\cite{garg2022can,hollmann2022tabpfn,von2023transformers}, which has been recently proved to be effective in reinforcement learning and bandit problems by methods like Decision Pretrained Transformers (DPT)~\citep{lee2023supervisedpretraininglearnincontext}. We call our model \method\ or \methodabbrv\ for short. \methodabbrv\ is two-fold: (i) During the \emph{offline} phase, we employ a history-dependent loss function to train a \emph{single} policy model that predicts the preferred responses given the history of responses within each preference criterion. In particular, we follow a direct preference optimization (DPO) approach to avoid learning a separate reward model~\citep{rafailov2023direct}. (ii) During the \emph{online} inference phase, for each new user, we follow an in-context learning approach by generating two potential responses for each prompt the user gives and asking the user to rank them. We then append those interactions to the trained model's context and continue the inference. This procedure allows the model to dynamically adapt to individual user preferences as the interaction progresses, rather than relying on a distinct validation set for model selection, as in prior work. See \cref{fig:main-figure} for an illustration of \methodabbrv.

We demonstrate the effectiveness of our approach in a contextual bandit setting, showcasing its ability to achieve personalized in-context learning by outperforming the Personalized Soups baseline while reducing the computation cost by only learning one policy model. Our results suggest the potential of in-context learning for scalable and efficient personalization in LLMs.

\captionsetup{font=footnotesize,labelfont=footnotesize}

\section{\method}
Consider a multi-preference learning setting, with a set of questions (prompts or contexts) $\cX$, potential responses $\cA$, and preference groups (or preference criteria) denoted by $\cG = \lcbar 1, 2, \ldots, K \rcbar$ for $K$ number of groups. For each group $g \in \cG$, assume an unknown function $r_g: \cX \times \cA \to \R$, which assigns a reward value $r_g\lpar a \given x \rpar$ to a potential response $a$ given a question $x$ and a preference group $g$. Moreover, assume there exist noisily rational annotators corresponding to different preference groups. An annotator from preference group $g$, prefers a response $a_w$ over $a_l$ for a question $x$ following a Bradley–Terry model \cite{christiano2017deep, bradley1952rank}:
\begin{align}
    \P \lpar a_w,a_l \given x \midsem g \rpar = \frac{\exp \lcbar r_g\lpar a_w \given x  \rpar \rcbar }{\exp \lcbar  r_g\lpar a_w \given x  \rpar \rcbar + \exp \lcbar  r_g\lpar {a}_l \given x\rpar \rcbar} ,
\end{align}
where in our notation, given an option between $a_w$ and $a_l$, $a_w$ was chosen by the user. Finally, we do not directly observe the reward values but only the offline preference data as a proxy:

\xhdr{Offline preference data} We assume the given offline datasets are stratified by the preference groups, i.e., 
\begin{align}
    \data_\pre = \lcbar \data^{(g)}_\pre \triangleq \lpar  x_{1}^{(g)}, a_{w,1}^{(g)}, {{a}}_{l,1}^{(g)}, x_{2}^{(g)}, a_{w,2}^{(g)}, {{a}}_{l,2}^{(g)}, \ldots,  x_{N}^{(g)}, a_{w,N}^{(g)}, {{a}}_{l,N}^{(g)} \rpar\rcbar_{g=1}^{K}.
\end{align}
In this notation, each superscript $^{(g)}$ corresponds to a fixed preference group $g \in \cG$, and each subscript $_{i}$ refers to a different question $x_i$ and its preference data $a_{w,i}, {a}_{l,i}$ for $N$ questions.~\footnote{The setting is similar to a contextual bandit rather than a Markov decision process.}

\xhdr{User preference profile} As discussed, we assume separate offline preference data for each preference group. However, we also allow for any preference profile induced by rewards in the convex hull of the true reward models of each preference group. Concretely, we denote a user preference profile by $z = \lpar \alpha_1, \alpha_2, \ldots, \alpha_K \rpar$ for $\alpha_1, \ldots, \alpha_K \geq 0$ and $\sum_{g=1}^K \alpha_g = 1$. Then, the corresponding reward function is defined as a linear combination of multiple reward functions, i.e., $r_z \lpar a \given x \rpar = \sum_{g=1}^K \alpha_i \cdot r_{g}\lpar a \given x \rpar$. We denote the set of all user preference profiles by $\Delta(\cG)$.

\xhdr{Online phase} Given the offline datasets $\data_\pre \in \mathcal{D}$, our goal is to train an \emph{in-context learning} model $M:  \cX \times \mathcal{D} \to \Delta\lpar \cA \rpar$ that can generate \emph{personalized} responses during the online deployment phase. The model interacts with a fixed user $z^\star \in \Delta\lpar \cG \rpar$, which is unknown to the model, for iterations $j=1, 2, \ldots$ as follows:
\begin{enumerate}
    \item The user (with profile $z^\star$) asks a question $x_j$ from the model.
    \item The model generates two potential responses given the question and the history of interactions, i.e., $a_{w,j}, a_{l,j} \sim M\lpar x_{j} \midsem x_1, a_{w,1}, {a}_{l,1}, \ldots, x_{j-1}, a_{w, j-1}, {a}_{l, j-1} \rpar$ or $a_{w,j}, a_{l,j} \sim  M\lpar x_j \rpar$ if $j = 1$. 
    \item The user prefers $a_{w,j}$ over ${a}_{l,j}$ based on $\P \lpar a_{w,j}, a_{l,j} \given x_j \midsem z^\star \rpar$.
\end{enumerate}

\xhdr{\method} Here, we present and motivate our proposed approach. We assume the questions in the offline and online datasets come from an unknown distribution $\dist_\cX$. Given a model $M_\theta$ and a preference profile $z^\star$, denote a history of interactions between the model and the user up to time $h$ as $\history_h = \lpar x_j, a_{w,j}, {a}_{l,j}\rpar_{j=1}^h$ with $\history_0 = \emptyset$. The joint distribution of the history is given as follows:
{\small
\begin{align}
    \dist_\cH\lpar \history_{h} \midsem M_\theta, z^\star \rpar = \prod_{j=1}^h \dist_\cX(x_j) \cdot M_\theta\lpar x_j \midsem \history_{j-1} \rpar(a_{w,j}) \cdot M_\theta\lpar x_j \midsem \history_{j-1} \rpar({a}_{l,j}) \cdot \P \lpar a_{w,j},{a}_{l,j} \given x_j \midsem z^\star \rpar.
\end{align}
}%
We define the notion of \emph{optimal in-context learning models} as the ones maximizing the following:
{
\begin{align}
    \max_{M_\theta}\, \E_{x \sim \dist_\cX, \history \sim \dist_\cH\lpar \cdot \midsem M_\theta, z^\star \rpar, a \sim M_\theta \lpar x \midsem \history \rpar} \lbar r_{z^\star}\lpar a \given x \rpar \rbar - \beta \cdot \kl{M_\theta\lpar x\midsem \history\rpar }{\pi_{\text{ref}}\lpar \cdot \given x \rpar}. \label{eq:optimal-model}
\end{align}
}%
This objective function trades off between exploration and exploitation by maximizing the reward of the generated responses while allowing the model full control over the previously generated answers in history. Moreover, $\beta$ is a parameter controlling the deviation from some reference policy $\pi_{\text{ref}}: \cX \to \Delta(\cA)$.
However, exact optimization of the objective function \cref{eq:optimal-model} is challenging since (i) the history (in-context data) $\history$ in the expectation also depends on the learnable model $M_\theta$, and (ii) $z^\star$ is unknown. Here, we propose an approximation to this objective in two ways. First, we replace the model $M_\theta$ with the reference policy $\pi_{\text{ref}}$ in the history distribution $\dist_\history$. Such approximation is reasonable since the Lagrangian in \cref{eq:optimal-model} already requires the learned model $M_\theta$ to be within a KL-ball from the reference policy. Second, we assume a uniform distribution $\texttt{Unif}(K)$ for $z^\star$ over different preference groups, i.e., $\P \lpar z^\star = g \rpar = \frac{1}{K}$ for all $g \in \cG$. While this assumption does not consider the users inside the reward convex hull, our empirical results demonstrate competitive performance for such users. Applying these two approximations, we get the following objective function:
{
\begin{align}
    \max_{M_\theta}\, &\E_{x \sim \dist_\cX, z^\star \sim \text{Unif}(K),  \history \sim \dist_\cH\lpar \cdot \midsem \pi_{\text{ref}}, z^\star \rpar, a \sim M_\theta \lpar x \midsem \history \rpar} \lbar r_{z^\star}\lpar a \given x \rpar \rbar - \beta \cdot \kl{M_\theta\lpar x\midsem \history\rpar }{\pi_{\text{ref}}\lpar \cdot \given x \rpar}. \label{eq:approx-model}
\end{align}
}%
Noting that $\data^{(g)}_\pre \sim \dist_\cH\lpar \cdot \midsem \pi_{\text{ref}}, g\rpar$, the above can be simplified as
{
\begin{align}
    \max_{M_\theta}\, \frac{1}{K} \sum_{g=1}^K \lpar \E_{x \sim \dist_\cX, \history \sim {\data^{(g)}_\pre}, a \sim M_\theta \lpar x \midsem \history \rpar} \lbar r_{g}\lpar a \given x \rpar \rbar  - \beta \kl{M_\theta\lpar x\midsem \history\rpar(a) }{\pi_{\text{ref}}\lpar a \given x \rpar} \rpar . \label{eq:approx-model}
\end{align}
}%
Instead of learning a reward model and optimizing the above objective, we follow a direct preference optimization (DPO) approach that results in the following loss function~\citep{rafailov2023direct}:
{\small
\begin{align}
    \loss\lpar \theta \rpar = - \frac{1}{K} \sum_{g=1}^K \E_{\lcbar \lpar x, a_w, {a_l} \rpar, H \rcbar \sim \data^{(g)}_\pre} \lbar \log \sigma \lpar {\beta} \log \frac{M_\theta\lpar x\midsem H \rpar(a_w)}{\pi_{\text{ref}}\lpar a_w \given x \rpar} - {\beta} \log \frac{M_\theta\lpar x\midsem H \rpar(a_l)}{\pi_{\text{ref}}\lpar {a_l} \given x \rpar}\rpar \rbar. \label{eq:loss_fn}
\end{align}
}%

\section{Experiments}

LLMs can be thought of as a contextual bandit, where the prompt can be represented as the context $x$. Then, each response corresponds to an arm of the bandit $a \in \mathcal{A}$. The reward for each response then is a function of both the prompt $x$ as well as the chosen arm $a$. For group $z$ let this reward function be represented as $r_z(a, x)$. The preferences between arms are generates through the Bradley-Terry model as described before \cite{christiano2017deep, bradley1952rank}. In this preliminary version, we use this framework to run proof-of-concept experiments for our algorithm.

\subsection{Experimental Setup}

We consider a simplified scenario involving three subpopulations, where each context vector $x$ represents a question and each action $a\in\{0, 1, 2, 3\}$ represents one of four possible responses. To create the preference dataset, we uniformly sample $N_c$ context vectors from the unit hypercube $[0, 1]^3$. All the contexts are annotated by users from the first subpopulation, 80\% are annotated by users from the second subpopulation, and 60\% by users from the third subpopulation. For the annotation process, we generate two candidate actions $a', a''$ from a uniformly random reference policy $\pi_\text{SFT} = (\frac{1}{4}, \frac{1}{4}, \frac{1}{4}, \frac{1}{4})$ and rank the two actions using the subpopulation-specific reward models. The reward model is formulated as a linear function, similar to the linear contextual bandit setup:
{\small
\begin{align}
r_z(a, x) = f_\phi(x)^\top\theta_z(a) + \cN \lpar 0, \sigma \rpar
\end{align}
}%
where $f_\phi: \mathbb{R}^3 \to \mathbb{R}^4$ is a context encoding function, $\theta_z$ is a reward matrix for subpopulation $z$ and $\mathcal{N}(0, \sigma)$ is a Gaussian noise with mean 0 and standard deviation $\sigma = 0.01$. We define $\theta_z$ such that each column $\theta_z(j)$ corresponds to the rewards obtained by selecting action $j$ for a user from group $z$. We have $\theta_i(j) = [\theta_{i1}(j), \theta_{i2}(j), \theta_{i3}(j), \theta_{i4}(j)]$, where $\theta_{i1}(j) = \theta_{i2}(j) = \theta_{i3}(j) = \theta_{i4}(j)=r_{ij}$ for any group $i$ and action $j$, we choose $r_{ij}\in\{1, 3, 5, 7\}$ to ensure that the reward model is deterministic and contrasting (i.e., there is a unique ranking of actions for each group.) We choose the context encoding function $f_\phi$ to be a linear function. We then train a simple transformer using history-dependent DPO on $\widetilde{\data}_\pre = \lcbar \lpar x_t, a_t, \bar{a}_t \rpar\rcbar_{j=1}^{T}$ where $T = 15$, with the loss function \cref{eq:loss_fn}. For our baseline Personalized Soup (PS) \cite{jang2023personalizedsoupspersonalizedlarge}, we trained 3 transformers, one for each subpopulation, using standard DPO on data specific to each subpopulation. In both cases, the transformers consist of 6 layers, 4 attention heads, and a hidden dimension of 256.

\subsection{Results}
We sample $L = 50$ test contexts $\{x^{\text{test}}_i\}_{i = 1}^{50}$ from the training set and evaluate the rewards obtained at each turn for $T = 15$ turns, for users from each group and for a user that has mixed reward function $r_{\tilde{z}}(a, x) = \pmb{\lambda}^\top r(a, x)$ where $\pmb{\lambda}\in\mathbb{R}^3$ is sampled randomly from the 3-simplex and $r(a, x) = [r_1(a, x), r_2(a, x), r_3(a, x)]$.

We evaluate PPT against the baseline which is Personalized Soups (PS) \cite{jang2023personalizedsoupspersonalizedlarge}. For PS, we begin by constructing 100 interpolated models by performing weighted linear interpolation of the three pretrained transformers. We initialize the best model $M^\star$ by randomly sampling one from 100 interpolated models. For each turn $t$, we evaluate the mean accuracy of $M^\star$ in predicting the preferred action given $\{x^{\text{test}}_i\}_{i = 1}^{L}$. We then sample a random context $x^{\text{val}}_j$ and generate two actions $a', a''$ from a uniform reference policy. Then we select the winning action denoted $a_w$ based on the test user's preferences. For each interpolated model, we compute the log-probability of $a_w$ for the context $x^\text{val}$ and add to the model's cumulative score. $M^\star$ is then updated based on the accumulated score. 

For our model, we maintain a history of interactions $\history$ for $T$ turns. We use the same set of validation contexts $\{x^{\text{val}}_j\}_{j = 1}^{T}$ and test contexts $\{x^{\text{test}}_i\}_{i = 1}^{L}$. For each turn $t$, we sample responses from $M_\theta (\cdot \vert  x^{\text{test}}_i, \history)$ and similarly evaluate the accuracy of $M_\theta$ in predicting the preferred action on $\{x^{\text{test}}_i\}_{i = 1}^{L}$. We then generate two actions $a', a''$ from $M_\theta (\cdot \vert  x^{\text{val}}_j, \history)$ and select $a_w$ based on the test user's preferences. The history is then updated with $\history \leftarrow \history \cup \{(x^{\text{val}}_j, a_j, \bar{a}_j)\}$.

\begin{figure}[htbp]
    \centering
    \begin{subfigure}[t]{\textwidth}
        \includegraphics[width=\textwidth]{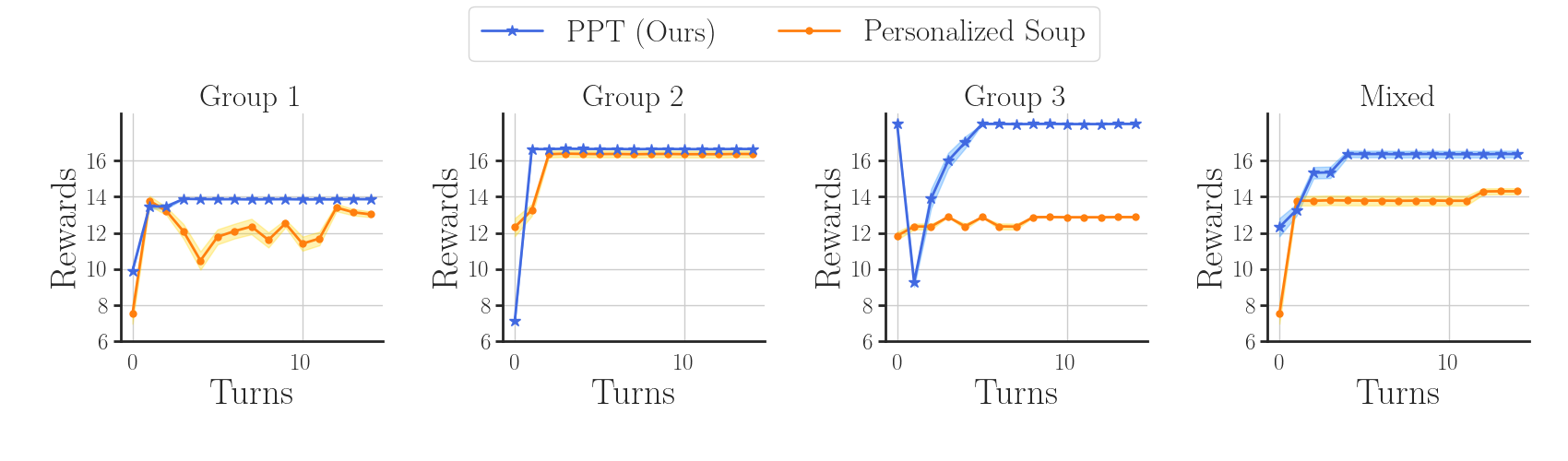}
        \vspace{-20pt}
        \caption{\small Rewards vs Turns ($N_c = 500$)}
        \label{fig:rewards_500}
    \end{subfigure}
    \begin{subfigure}[t]{\textwidth}
        \vspace{10pt}
        \includegraphics[width=\textwidth]{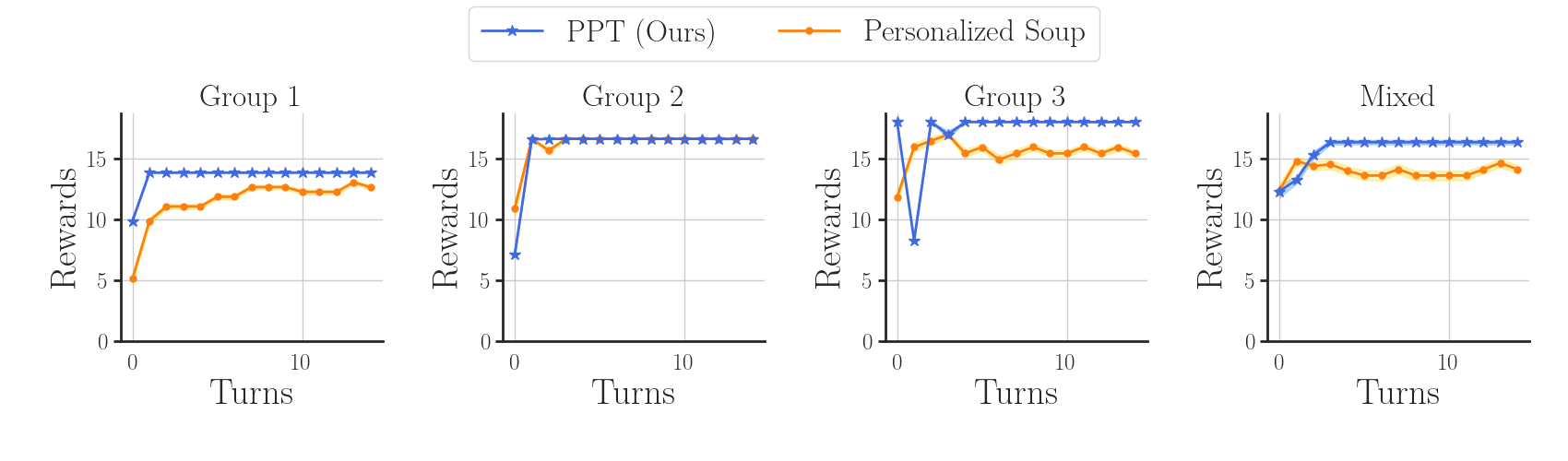}
        \vspace{-20pt}
        \caption{\small Rewards vs Turns ($N_c = 1000$)}
        \label{fig:rewards_1000}
    \end{subfigure}
    \caption{\small Comparison of rewards between PPT (ours) and the Personalized Soups (PS) over 15 interaction turns for different user groups. Figure~\ref{fig:rewards_500} and Figure~\ref{fig:rewards_1000} show results with $N_c = 500$ and $N_c = 1000$ context vectors, respectively. Each subplot corresponds to tests with users from one of the three subpopulations and a user with mixed preferences. The results demonstrate that PPT consistently outperforms the PS baseline across all groups. The increase in rewards for our method as the number of turns grows indicates effective in-context learning and dynamic adaptation to user preferences.}
    \vspace{-10pt}
    \label{fig:rewards}
\end{figure}

Figure~\ref{fig:rewards} shows that our method consistently outperforms PS across all user groups. The rewards increase with the number of turns. In contrast, PS shows some fluctuations in rewards even with the large number of interpolated models and turns. This highlights the advantage of our approach in not requiring separate models for each subpopulation; instead, we train a single model capable of adapting to any user preference. Furthermore, the consistent improvement in rewards across different groups, including users with mixed preferences, demonstrates the robustness of our method. As the number of turns grows, our model becomes increasingly accurate in predicting the user's preferred actions, even if it does not perform optimally in the initial iterations. This behavior underscores \methodabbrv's ability to effectively learn in-context, dynamically adapting without the need for retraining or complex model selection procedures.

\section{Conclusion}

In this paper, we introduced the Preference Pretrained Transformer (PPT), a novel framework for online personalized adaptation of language models to user preferences. PPT addresses the limitations of existing RLHF approaches by leveraging the in-context learning capabilities of transformers to dynamically adapt to individual user preferences. We demonstrated the effectiveness of PPT in a contextual bandit setting, showing that our model achieves personalized adaptation superior to the Personalized Soups baseline while significantly reducing computational costs. By training a single policy model instead of multiple preference-specific models, PPT offers a more scalable and efficient solution for LM personalization.
A key limitation of this work is the lack of experiments with large language models, which will be crucial for validating the approach's scalability and real-world applicability. Future work should explore the application of PPT to more complex language tasks and investigate its performance with larger language models and diverse user populations. 

\bibliography{main}

\begin{thebibliography}{20}
\providecommand{\natexlab}[1]{#1}
\providecommand{\url}[1]{\texttt{#1}}
\expandafter\ifx\csname urlstyle\endcsname\relax
  \providecommand{\doi}[1]{doi: #1}\else
  \providecommand{\doi}{doi: \begingroup \urlstyle{rm}\Url}\fi

\bibitem[Stiennon et~al.(2020)Stiennon, Ouyang, Wu, Ziegler, Lowe, Voss,
  Radford, Amodei, and Christiano]{stiennon2020learning}
Nisan Stiennon, Long Ouyang, Jeffrey Wu, Daniel Ziegler, Ryan Lowe, Chelsea
  Voss, Alec Radford, Dario Amodei, and Paul~F Christiano.
\newblock Learning to summarize with human feedback.
\newblock \emph{Advances in Neural Information Processing Systems},
  33:\penalty0 3008--3021, 2020.

\bibitem[Nakano et~al.(2021)Nakano, Hilton, Balaji, Wu, Ouyang, Kim, Hesse,
  Jain, Kosaraju, Saunders, et~al.]{nakano2021webgpt}
Reiichiro Nakano, Jacob Hilton, Suchir Balaji, Jeff Wu, Long Ouyang, Christina
  Kim, Christopher Hesse, Shantanu Jain, Vineet Kosaraju, William Saunders,
  et~al.
\newblock Webgpt: Browser-assisted question-answering with human feedback.
\newblock \emph{arXiv preprint arXiv:2112.09332}, 2021.

\bibitem[Bai et~al.(2022)Bai, Jones, Ndousse, Askell, Chen, DasSarma, Drain,
  Fort, Ganguli, Henighan, et~al.]{bai2022training}
Yuntao Bai, Andy Jones, Kamal Ndousse, Amanda Askell, Anna Chen, Nova DasSarma,
  Dawn Drain, Stanislav Fort, Deep Ganguli, Tom Henighan, et~al.
\newblock Training a helpful and harmless assistant with reinforcement learning
  from human feedback.
\newblock \emph{arXiv preprint arXiv:2204.05862}, 2022.

\bibitem[Ouyang et~al.(2022)Ouyang, Wu, Jiang, Almeida, Wainwright, Mishkin,
  Zhang, Agarwal, Slama, Ray, et~al.]{ouyang2022training}
Long Ouyang, Jeffrey Wu, Xu~Jiang, Diogo Almeida, Carroll Wainwright, Pamela
  Mishkin, Chong Zhang, Sandhini Agarwal, Katarina Slama, Alex Ray, et~al.
\newblock Training language models to follow instructions with human feedback.
\newblock \emph{Advances in neural information processing systems},
  35:\penalty0 27730--27744, 2022.

\bibitem[Vamplew et~al.(2018)Vamplew, Dazeley, Foale, Firmin, and
  Mummery]{vamplew2018human}
Peter Vamplew, Richard Dazeley, Cameron Foale, Sally Firmin, and Jane Mummery.
\newblock Human-aligned artificial intelligence is a multiobjective problem.
\newblock \emph{Ethics and information technology}, 20:\penalty0 27--40, 2018.

\bibitem[Ramé et~al.(2023)Ramé, Couairon, Shukor, Dancette, Gaya, Soulier,
  and Cord]{rame2023rewardedsoupsparetooptimalalignment}
Alexandre Ramé, Guillaume Couairon, Mustafa Shukor, Corentin Dancette,
  Jean-Baptiste Gaya, Laure Soulier, and Matthieu Cord.
\newblock Rewarded soups: towards pareto-optimal alignment by interpolating
  weights fine-tuned on diverse rewards, 2023.
\newblock URL \url{https://arxiv.org/abs/2306.04488}.

\bibitem[Jang et~al.(2023)Jang, Kim, Lin, Wang, Hessel, Zettlemoyer,
  Hajishirzi, Choi, and
  Ammanabrolu]{jang2023personalizedsoupspersonalizedlarge}
Joel Jang, Seungone Kim, Bill~Yuchen Lin, Yizhong Wang, Jack Hessel, Luke
  Zettlemoyer, Hannaneh Hajishirzi, Yejin Choi, and Prithviraj Ammanabrolu.
\newblock Personalized soups: Personalized large language model alignment via
  post-hoc parameter merging, 2023.
\newblock URL \url{https://arxiv.org/abs/2310.11564}.

\bibitem[Casper et~al.(2023)Casper, Davies, Shi, Gilbert, Scheurer, Rando,
  Freedman, Korbak, Lindner, Freire, et~al.]{casper2023open}
Stephen Casper, Xander Davies, Claudia Shi, Thomas~Krendl Gilbert,
  J{\'e}r{\'e}my Scheurer, Javier Rando, Rachel Freedman, Tomasz Korbak, David
  Lindner, Pedro Freire, et~al.
\newblock Open problems and fundamental limitations of reinforcement learning
  from human feedback.
\newblock \emph{arXiv preprint arXiv:2307.15217}, 2023.

\bibitem[Chakraborty et~al.(2024)Chakraborty, Qiu, Yuan, Koppel, Huang,
  Manocha, Bedi, and Wang]{chakraborty2024maxminrlhfequitablealignmentlarge}
Souradip Chakraborty, Jiahao Qiu, Hui Yuan, Alec Koppel, Furong Huang, Dinesh
  Manocha, Amrit~Singh Bedi, and Mengdi Wang.
\newblock Maxmin-rlhf: Towards equitable alignment of large language models
  with diverse human preferences, 2024.
\newblock URL \url{https://arxiv.org/abs/2402.08925}.

\bibitem[Barrett and Narayanan(2008)]{barrett2008learning}
Leon Barrett and Srini Narayanan.
\newblock Learning all optimal policies with multiple criteria.
\newblock In \emph{Proceedings of the 25th international conference on Machine
  learning}, pages 41--47, 2008.

\bibitem[Li et~al.(2020)Li, Zhang, and Wang]{li2020deep}
Kaiwen Li, Tao Zhang, and Rui Wang.
\newblock Deep reinforcement learning for multiobjective optimization.
\newblock \emph{IEEE transactions on cybernetics}, 51\penalty0 (6):\penalty0
  3103--3114, 2020.

\bibitem[Marta et~al.(2023)Marta, Holk, Pek, Tumova, and
  Leite]{marta2023aligning}
Daniel Marta, Simon Holk, Christian Pek, Jana Tumova, and Iolanda Leite.
\newblock Aligning human preferences with baseline objectives in reinforcement
  learning.
\newblock In \emph{2023 IEEE International Conference on Robotics and
  Automation (ICRA)}, pages 7562--7568. IEEE, 2023.

\bibitem[Wu et~al.(2024)Wu, Hu, Shi, Dziri, Suhr, Ammanabrolu, Smith,
  Ostendorf, and Hajishirzi]{wu2024fine}
Zeqiu Wu, Yushi Hu, Weijia Shi, Nouha Dziri, Alane Suhr, Prithviraj
  Ammanabrolu, Noah~A Smith, Mari Ostendorf, and Hannaneh Hajishirzi.
\newblock Fine-grained human feedback gives better rewards for language model
  training.
\newblock \emph{Advances in Neural Information Processing Systems}, 36, 2024.

\bibitem[Garg et~al.(2022)Garg, Tsipras, Liang, and Valiant]{garg2022can}
Shivam Garg, Dimitris Tsipras, Percy~S Liang, and Gregory Valiant.
\newblock What can transformers learn in-context? a case study of simple
  function classes.
\newblock \emph{Advances in Neural Information Processing Systems},
  35:\penalty0 30583--30598, 2022.

\bibitem[Hollmann et~al.(2022)Hollmann, M{\"u}ller, Eggensperger, and
  Hutter]{hollmann2022tabpfn}
Noah Hollmann, Samuel M{\"u}ller, Katharina Eggensperger, and Frank Hutter.
\newblock Tabpfn: A transformer that solves small tabular classification
  problems in a second.
\newblock \emph{arXiv preprint arXiv:2207.01848}, 2022.

\bibitem[Von~Oswald et~al.(2023)Von~Oswald, Niklasson, Randazzo, Sacramento,
  Mordvintsev, Zhmoginov, and Vladymyrov]{von2023transformers}
Johannes Von~Oswald, Eyvind Niklasson, Ettore Randazzo, Jo{\~a}o Sacramento,
  Alexander Mordvintsev, Andrey Zhmoginov, and Max Vladymyrov.
\newblock Transformers learn in-context by gradient descent.
\newblock In \emph{International Conference on Machine Learning}, pages
  35151--35174. PMLR, 2023.

\bibitem[Lee et~al.(2023)Lee, Xie, Pacchiano, Chandak, Finn, Nachum, and
  Brunskill]{lee2023supervisedpretraininglearnincontext}
Jonathan~N. Lee, Annie Xie, Aldo Pacchiano, Yash Chandak, Chelsea Finn, Ofir
  Nachum, and Emma Brunskill.
\newblock Supervised pretraining can learn in-context reinforcement learning,
  2023.
\newblock URL \url{https://arxiv.org/abs/2306.14892}.

\bibitem[Rafailov et~al.(2023)Rafailov, Sharma, Mitchell, Manning, Ermon, and
  Finn]{rafailov2023direct}
Rafael Rafailov, Archit Sharma, Eric Mitchell, Christopher~D Manning, Stefano
  Ermon, and Chelsea Finn.
\newblock Direct preference optimization: Your language model is secretly a
  reward model.
\newblock In \emph{Thirty-seventh Conference on Neural Information Processing
  Systems}, 2023.

\bibitem[Christiano et~al.(2017)Christiano, Leike, Brown, Martic, Legg, and
  Amodei]{christiano2017deep}
Paul~F Christiano, Jan Leike, Tom Brown, Miljan Martic, Shane Legg, and Dario
  Amodei.
\newblock Deep reinforcement learning from human preferences.
\newblock \emph{Advances in neural information processing systems}, 30, 2017.

\bibitem[Bradley and Terry(1952)]{bradley1952rank}
Ralph~Allan Bradley and Milton~E Terry.
\newblock Rank analysis of incomplete block designs: I. the method of paired
  comparisons.
\newblock \emph{Biometrika}, 39\penalty0 (3/4):\penalty0 324--345, 1952.

\end{thebibliography}
\end{document}